\pdfoutput=1
\documentclass[11pt]{article}

\usepackage[preprint]{acl}

\usepackage{times}
\usepackage{latexsym}

\usepackage[T1]{fontenc}

\usepackage[utf8]{inputenc}

\usepackage{microtype}

\usepackage{inconsolata}

\usepackage{graphicx}

\title{Decoding with Limited Teacher Supervision Requires\\Understanding When to Trust the Teacher}

\definecolor{brickred}{rgb}{0.6, 0, 0}

\usepackage{booktabs}
\usepackage{graphicx} 
\usepackage{multirow} 
\usepackage{caption}
\usepackage{algorithm}
\usepackage{algorithmic}
\usepackage{amsmath,amssymb,amsfonts}
\usepackage{array}
\usepackage{tabu}
\usepackage{verbatim}
\usepackage{listings}
\usepackage{cleveref}

\usepackage{xspace}
\makeatletter
\DeclareRobustCommand\onedot{\futurelet\@let@token\@onedot}
\def\@onedot{\ifx\@let@token.\else.\null\fi\xspace}
\def\eg{\emph{e.g}\onedot} 
\def\ie{\emph{i.e}\onedot}

\makeatother

\author{Hyunjong Ok, Jegwang Ryu, Jaeho Lee\thanks{\hspace{1mm} corresponding author}\\
Pohang University of Science and Technology (POSTECH) \\
\texttt{hyunjong.ok@gmail.com, \{jegwang.ryu, jaeho.lee\}@postech.ac.kr}
 }

\begin{document}
\maketitle
\begin{abstract}
How can small-scale large language models (LLMs) efficiently utilize the supervision of LLMs to improve their generative quality? This question has been well studied in scenarios where there is no restriction on the number of LLM supervisions one can use, giving birth to many decoding algorithms that utilize supervision without further training. However, it is still unclear what is an effective strategy under the \textit{limited supervision} scenario, where we assume that no more than a few tokens can be generated by LLMs. To this end, we develop an algorithm to effectively aggregate the small-scale LLM and LLM predictions on initial tokens so that the generated tokens can more accurately condition the subsequent token generation by small-scale LLM only. Critically, we find that it is essential to adaptively overtrust or disregard the LLM prediction based on the confidence of the small-scale LLM. Through our experiments on a wide range of models and datasets, we demonstrate that our method provides a consistent improvement over conventional decoding strategies.\\
{\small \textbf{Code:} \url{https://github.com/HJ-Ok/DecLimSup}}
\end{abstract}

\begin{figure*}[t]
    \centering
    \includegraphics[width=2 \columnwidth]{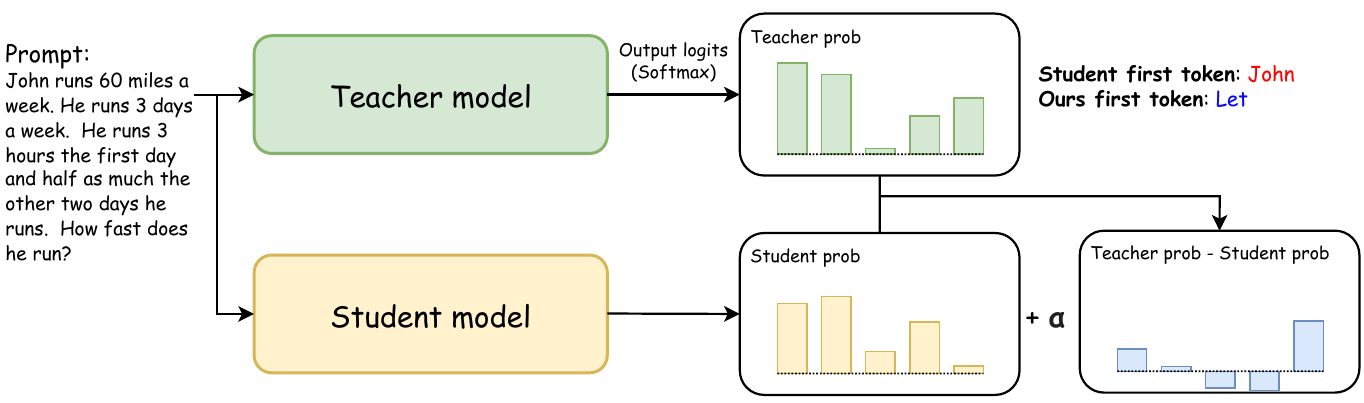}
    
    \caption{Illustration of our methodology. As the parameter $\alpha$ increases, the methodology leverages a more significant disparity in knowledge between the teacher and the student models. The example shows a modification in the initial generated word from `John' to `Let,' which allows the sentence to generate the correct answer when subsequently generated with the student model.}
    \label{fig:TUNA_overall_method}
    \vspace{-1em}
\end{figure*}

\begin{figure*}[t]
    \centering
    \includegraphics[width=2 \columnwidth]{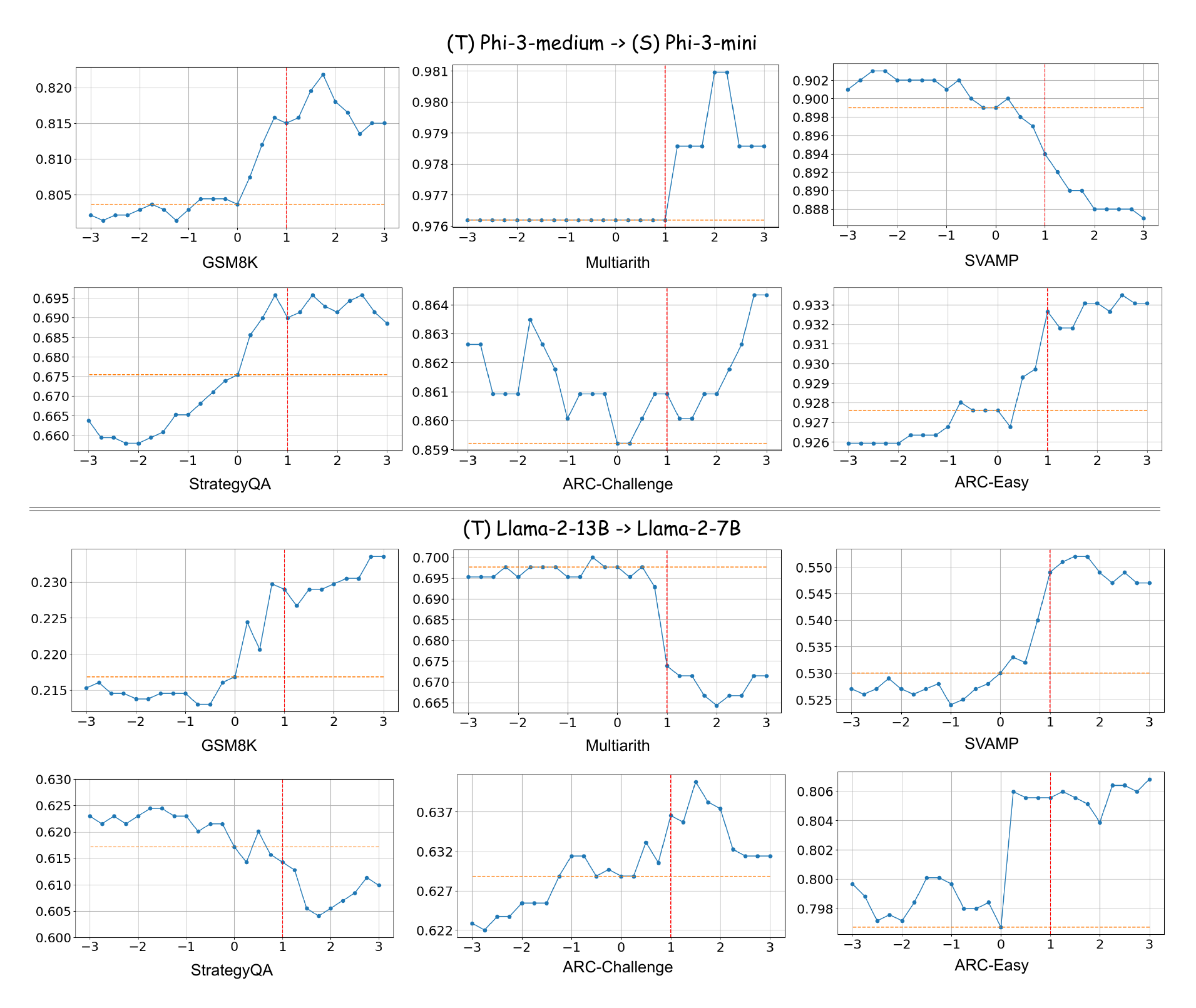}
    
    \caption{Visualization of accuracy as a function of $\alpha$. The red dashed line indicates $\alpha$=1, and the orange line represents the student model's baseline performance.}
    \label{fig:TUNA_alpha_ablation}
\end{figure*}

\section{Introduction}\label{sec:intro}

Large language models (LLMs) have demonstrated their tremendous capability to generate human-like text sentences that convey rich knowledge in various problem domains \citep{achiam2023gpt,bai2022training,team2023gemini,touvron2023llama}. However, due to their gigantic model scale and autoregressive nature based on next-token generation, LLMs often suffer from having a significantly high latency \citep{xu23}.

Small-scale LLMs, called sLLMs, have thus garnered much attention \citep{gunasekar2023textbooks, li2023textbooks, abdin2024phi}. sLLMs can run much faster than LLMs, making them a promising alternative to LLMs for applications that require agile responses or on-device inference. However, sLLMs tend to perform clearly worse than their larger counterparts, especially for tasks that require in-depth reasoning \citep{biderman23}.

If we have both LLM and sLLM available, can we use both models to enjoy the quality of LLM and the speed of sLLM? A recent line of works shows that this is possible, even without any further training, by having two models collaborate on the decoding procedure: In speculative decoding \citep{leviathan23,xia23}, the candidate text is generated rapidly by the sLLM, which is then verified by the LLM to ensure its correctness. This method admits parallel inference, enabling a rapid generation of LLM-level responses. However, this approach requires a lot of memory to load multiple instances of LLMs and sLLMs, and thus, it is very difficult to use on edge devices.

In this paper, we consider an alternative decoding scenario with \textit{limited LLM supervision}. That is, we assume that one primarily decodes with sLLM (the \text{student}) but can utilize a very limited number of supervisions from the LLM (the \text{teacher}). The supervision may be as scarce as generating a single token, incorporating the scenario where the LLM is remotely located, \eg, at a central server. Critically, we no longer restrict the model to generate the exact same outcome as the LLM, but simply aim to maximize the generative quality itself.

A natural approach is to use two models to predict the same token and aggregate their predictions to generate a better token. Such an approach has been recently studied by \citet{li-etal-2023-contrastive}, without any limitation on the number of LLM supervision. Here, it turns out that an effective strategy is to \textit{overtrust} the teacher and negatively account for the student. Precisely, one takes a weighted sum of teacher and student logits with weights $1+\alpha$ and $-\alpha$, respectively. It has been observed that a single positive $\alpha=0.5$ works consistently well over diverse scenarios \citep{o2023contrastive}.

Under the limited supervision scenario, however, we make a critical observation that overtrusting the teacher no longer continues to be the dominant strategy. Intriguingly, we find that overtrusting the student works much better under certain setups, enjoying a better alignment with subsequently generated tokens where no supervision is available. In fact, our empirical analysis reveals that \textit{who should we overtrust by how much} highly varies over the choice of models and tasks, and even for each datum. In other words, for supervision-limited scenarios, we are in desperate need of a mechanism to determine whom to overtrust by how much.

To this end, we develop an algorithm to utilize the LLM supervision in an adaptive manner to improve sLLM performance. In particular, we identify that the entropy (\ie, confidence) of the generated tokens is highly correlated with whether one should trust the teacher or the student.
If the student's generated token has high or low entropy, overtrusting the student works better. This implies if the student is confident, the teacher's information could be the noise, and if the student ponders generating a token (\ie, high entropy), the teacher's information can cause confusion. Our method predicts the optimal $\alpha$ on a per-datum basis, consistently improving the predictive performance of sLLM over a wide range of models and tasks.

Our key contributions are threefold:

\begin{itemize}
\item We formulate and initiate research toward sLLM decoding with limited LLM supervision, which bears much practical importance.

\item We discover that, with limited supervision, the conventional strategies of overtrusting LLMs are largely suboptimal.

\item We propose a novel entropy-based mechanism to determine who to overtrust by how much among sLLM and LLM and demonstrate its effectiveness on a wide range of setups.

\end{itemize}
\section{Framework}\label{sec:framework}
We now formally describe the problem of decoding with limited teacher supervision and the overtrust-based framework to aggregate teacher and student predictions. Similar to recent works \citep{leviathan23,li-etal-2023-contrastive} and unlike distillation \citep{hinton15}, we do not assume that we can train the sLLM further.

We consider a setup where we have two models available: The teacher LLM and the student sLLM. Given some input prompt, we assume that we can invoke the teacher up to $N$ times and the student unlimited times to generate subsequent tokens.

For decoding, we consider aggregating the predictions of the teacher and student at the token-level. Concretely, let $f_{s}(x), f_{t}(x)$ be the prediction logits of the student and teacher. Then, we consider a prediction based on the aggregated softmax
\begin{align}
S_{\alpha} = \sigma(f_{s}(x)) + \alpha (\sigma(f_{t}(x)) - \sigma(f_{s}(x)))\label{equation1}
\end{align}
where $\sigma$ denotes the softmax function and $\alpha \in \mathbb{R}$ is a tunable parameter that determines which model should be trusted. If $\alpha = 1$, we are following the teacher's prediction, and if $\alpha = 0$, we are using the student prediction. Contrastive decoding \citep{li-etal-2023-contrastive} uses $\alpha > 1$, which is to overtrust the teacher and disregard the student.\footnote{Note that the contrastive decoding actually combines logits instead of softmax. We use softmax for our cases since it empirically works better.} In any case, the token is generated as an output achieving the maximum aggregated softmax $S_{\alpha}$ (Fig.~\ref{fig:TUNA_overall_method}).

As the number of teacher supervisions is limited, we make combined predictions (eq.~\ref{equation1}) only for the first $N$ tokens. That is, the text consists of
\begin{align}
(\underbrace{p_1,\cdots,p_m}_{\text{given prompt}},\underbrace{t_1,\cdots,t_N}_{\text{LLM + sLLM}},\underbrace{t_{N+1},\cdots}_{\text{sLLM only}}).
\end{align}
\section{Method}\label{sec:method}

Given this framework, we identify two core algorithmic questions. First, what is the optimal value of $\alpha$? Second, given some prompts, how can we determine whether we should use the teacher's knowledge or not, specifically for the datum?
 
\begin{figure*}[t]
    \centering
    \includegraphics[width=2 \columnwidth]{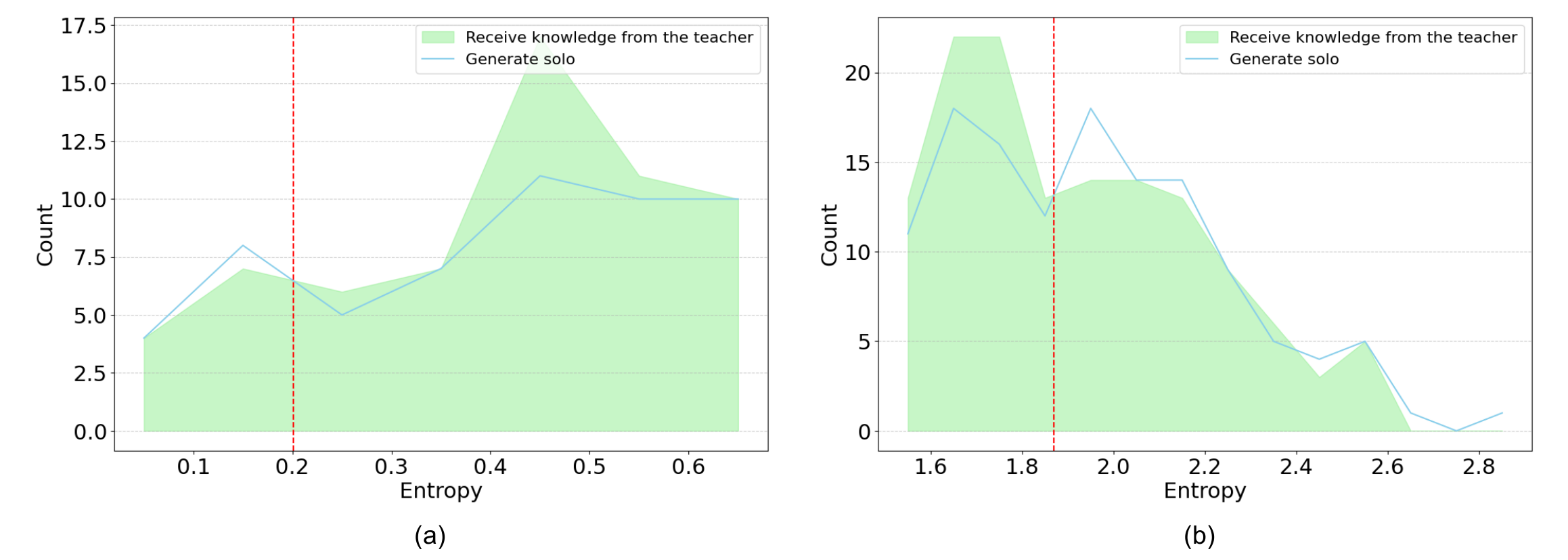}
    
    \caption{Visualization of the number of correct answers performed by ``Receive knowledge from the teacher'' and ``Generate solo'' based on student entropy values utilizing Llama-2 on GSM8K. (a) represents cases with low entropy, while (b) shows cases with high entropy. The red dashed line indicates the threshold beyond which ``Generate solo'' demonstrates superior performance.}
    \label{fig:two_cases}
\end{figure*}

\subsection{Key Observations} \label{subsec:entropy_method}

We now conduct a systematic empirical analysis to answer these questions.

\paragraph{(1) Strange case of $N=1$.} Under a wide range of setups, we have empirically evaluated the case of $N=1$ with varying \textit{trust hyperparameter} $\alpha$ (Fig.~\ref{fig:TUNA_alpha_ablation}). We make two intriguing observations.
\begin{itemize}
\item Even a \textit{single-token supervision} from LLM can \textit{\textbf{boost the accuracy substantially}}; in Phi3-mini evaluated on the StrategyQA dataset, the accuracy increases by $2\%\mathrm{p}$.
\item The \textit{optimal $\alpha$ significantly differs} from task to task and model to model. Surprisingly, there are certain cases where the optimal $\alpha$ is smaller than zero, \ie, overtrusting the student works better. Detailed case studies are in Appendix~\ref{detailed empirical study in entropy}. This contrasts with the case of unlimited supervision, where $\alpha=1.5$ works consistently well \citep{o2023contrastive}.
\end{itemize}

Summing up these observations, we conclude that we are in need of a good mechanism to predict the optimal $\alpha$ rather than using a fixed value.

\paragraph{(2) Entropy and supervision.} We also analyze the relationship between the prediction entropy of the student and the effect of teacher supervision (Fig.~\ref{fig:two_cases}), where we use the optimal $\alpha$ approximated from the previous analysis.

We observe that there exists certain \textit{interval of entropy} that whenever the student prediction entropy lies inside the interval, aggregating teacher predictions are likely to boost performance. If the student is extremely confident, the teacher's prediction will only add noise, leading to a lower accuracy. If the student is extremely unsure, aggregating teacher knowledge may disrupt the student's careful consideration and degrade the performance.


\subsection{Algorithmic implications} 
\label{subsec:alpha_model}

Motivated by these observations, we develop two mechanisms to best incorporate the limited supervision from the teacher LLM.

\begin{figure}[t]
    \centering
    \includegraphics[width=1 \columnwidth]{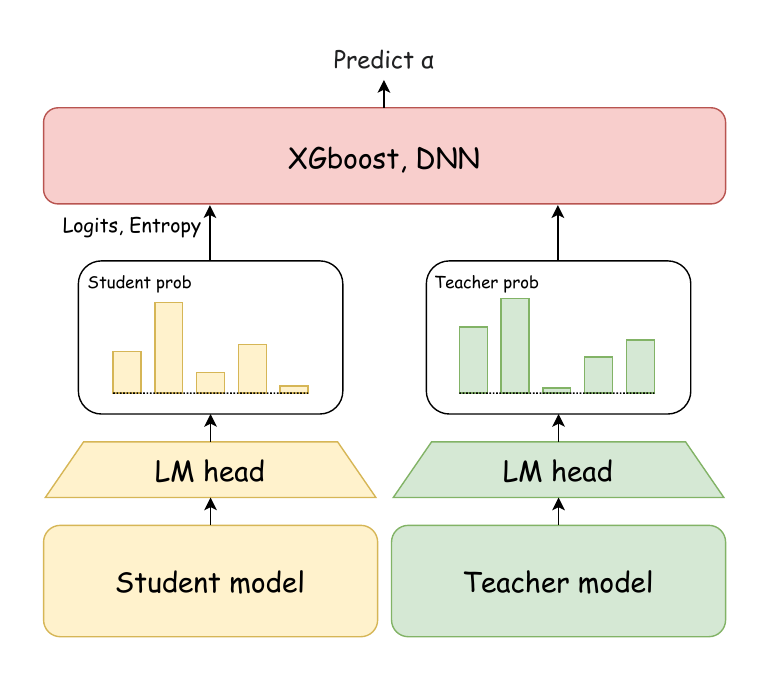}
    
    \caption{Illustration of the optimal $\alpha$ predict module.}
    \label{fig:TUNA_get_reference_method}
\end{figure}

\paragraph{Predicting optimal $\alpha$ for each datum.} To avoid performing an extensive search for the optimal $\alpha$ for each data point, we propose to train a predictor that estimates the optimal $\alpha$. More specifically, the model predicts $\alpha$ based on the teacher and student logits given as inputs (Fig.~\ref{fig:TUNA_get_reference_method}). We compare two different models, XGBoost and DNN.

\paragraph{Entropy-based knowledge injection.} We determine whether to utilize the teacher supervision or not based on an interval-based classifier using the student classification entropy (Algorithm~\ref{algorithm_1}).

\begin{algorithm}[h]
\caption{Entropy-based knowledge injection}
\begin{algorithmic}
\REQUIRE Entropy of the generated token $E$, two thresholds $T_1$ and $T_2$ ($T_1 < T_2$)

\IF{$T_1 < E < T_2$}
    \STATE Receive knowledge from the teacher
\ELSIF{$E \leq T_1$ or $T_2 \leq E$}
    \STATE Generate solo
\ENDIF
\end{algorithmic}
\label{algorithm_1}
\end{algorithm}

\section{Experimental Setup}
To demonstrate the superiority of our method, we evaluate our method on two types of tasks: classification tasks over multiple domains (\cref{ssec:classification}) and various LLM benchmarks (\cref{ssec:benchmark}). In \cref{ssec:method_experiment}, we describe the implementation details of our algorithm. Any detail that does not appear in this section is given in Appendix~\ref{app:details}.

\subsection{Classification Task}\label{ssec:classification}

For the classification task, we first fine-tuned each student and teacher model. Then, we predict the class based on this equation~\ref{equation1} in inference time.

\paragraph{Datasets.} For the classification task, we use three datasets: For image, we use the CIFAR-100~\cite{krizhevsky2009learning}; for audio, we use the ESC-50~\cite{piczak2015esc}; for text, we use the MNLI~\cite{williams2018broad} dataset.

\paragraph{Models.} For CIFAR-100, we use the DeiT~\cite{touvron2021training}. For ESC-50, we use the AST~\cite{gong21b_interspeech}. For MNLI, we use the DeBERTa~\cite{he2020deberta}. Details about teacher and student models exist in Appendix \ref{detail model}.

\paragraph{Other details.} For CIFAR-100 and MNLI, we report the average score from 5 runs with different random seeds for each setting. Experiments are done on both Intel Gaudi 2 or NVIDIA A6000, and we report the cross-validation score for ESC-50 following \citet{chen2022hts}.

\subsection{LLM Benchmark}\label{ssec:benchmark}

\paragraph{Datasets.} For LLM benchmarks, we evaluate a total of six different tasks. In particular, we use the GSM8K~\cite{cobbe2021training}, Multiarith~\cite{koncel-kedziorski-etal-2016-mawps}, SVAMP~\cite{patel2021nlp}, partition of MATH (MATH-Easy), ARC~\cite{clark2018think}, and StrategyQA~\cite{geva2021did} datasets.  Further details are available in the Appendix~\ref{detail data information}.

\paragraph{Models.} We have tested over a total of four different families of models: Phi-3~\cite{abdin2024phi}, Llama-2~\cite{touvron2023llama}, Vicuna~\cite{zheng2024judging}, and Qwen~\cite{qwen}. Detailed information on teacher and student models exists in Appendix \ref{detail model}.

\paragraph{Decoding.} To confirm the exact performance measurement, we do not apply sampling algorithms~\cite{fan-etal-2018-hierarchical, holtzman2019curious} during the decoding process.

\paragraph{Prompt.} We have provided few-shot prompts, with the number of shots ranging from 2 to 8. The prompts end with the phrase `the answer is.' Then, we extract that span and compare it with the ground truth answer.
More detailed information about the prompts is given in Appendix~\ref{full prompts}.

\paragraph{Hardware.} The experiments have mainly taken place on the Intel Gaudi 2, NVIDIA A100, and H100. The results have been cross-validated over devices. The model inference has been performed using the bfloat16 format.

\paragraph{Other details.} We search for the optimal $\alpha$ by exploring the range from 3 to -3 with 0.25 intervals. Detailed information about the evaluation strategy of each benchmark is in Appendix~\ref{LLM_benchmark_details}.

\subsection{Implementing our Algorithm}\label{ssec:method_experiment}

Our model, designed to predict the optimal $\alpha$, is initially conducted in inference across $\alpha$ ranging from 3 to -1 in intervals of 0.25 to make the training dataset. The dataset is constructed by labeling $\alpha$s that correctly predict the answer as one and those that do not as zero. Then, we train the model to perform multi-label binary classification. Our implementation uses logits from both student and teacher models as inputs. Additionally, we have incorporated the entropy of each logit as further input. The $\alpha$ with the highest confidence is selected as the output during inference. 

To demonstrate the superiority of our methodology, we mainly experiment with Phi-3, which shows the best performance on the GSM8K dataset. Since GSM8K is absent in the validation set, we split the training set into five cross-validation folds. Experiments are done on a single GeForce RTX 4090 GPU, and we conduct experiments using random seeds and report the average test score across these five folds.

\section{Results}
\subsection{Classification task} 
\begin{table}[t]
\centering
\vspace{0.25cm}
\resizebox{1\linewidth}{!}{
\begin{tabular}{lcccc}
\toprule
Methods & CIFAR-100 & ESC-50 & MNLI-m & MNLI-mm \\ \midrule
Student & 82.71 {$\pm$ 0.24} & 77.80 {$\pm$ 2.75} & 90.96 {$\pm$ 0.07}& 90.84 {$\pm$ 0.11} \\ 
Teacher & 89.47 {$\pm$ 0.16} & 94.55 {$\pm$ 0.84} & 91.60 {$\pm$ 0.06} & 91.58 {$\pm$ 0.13} \\ 
Ours & \textbf{89.75} {$\pm$ 0.15} & \textbf{94.75} {$\pm$ 0.98} & \textbf{91.91} {$\pm$ 0.10} & \textbf{91.99} {$\pm$ 0.19} \\ 
\bottomrule
\end{tabular}
}\vspace{-0.5em}
\caption{Experimental results on the classification task} \vspace{-1em}
\label{result_classification}
\end{table}
\begin{table}[t]
\centering
\vspace{0.25cm}
\resizebox{1\linewidth}{!}{
\begin{tabular}{lcccc}
\toprule
Student Model & Student acc. & Teacher acc. & Ours acc. & Optimal $\alpha$ \\ \midrule
Distill-BERT & $82.08 \pm 0.15$ & $91.60 \pm 0.06$ & $91.75 \pm 0.04$ & $0.75 \pm 0.13$ \\ 
DeBERTa-v3-small & $87.76 \pm 0.11$ & $91.60 \pm 0.06$ & $91.81 \pm 0.12$ & $0.73 \pm 0.12$ \\ 
DeBERTa-large & $90.96 \pm 0.07$ & $91.60 \pm 0.06$ & $91.91 \pm 0.10$ & $0.54 \pm 0.03$ \\
\bottomrule
\end{tabular}
}\vspace{-0.5em}
\caption{Experimental results on different students for optimal $\alpha$ ablation study on MNLI matched dataset.} \vspace{-1em}
\label{mnli_ablation}
\end{table}
\begin{table*}[!]
\small
\centering
\vspace{0.25cm}
\resizebox{1\textwidth}{!}{
\begin{tabular}{lccccccc}
\midrule
 Models & GSM8K & Multiarith & SVAMP & MATH-Easy & StrategyQA & ARC-Challenge & ARC-Easy\\ \midrule
Phi3-mini & 80.36 & 97.62 & 89.90 & 63.51 &  67.54 & 85.92 & 92.76 \\ 
\quad w/ reference ($\alpha$ = 1) & 81.50 & 97.62 & 89.40 & 64.39 & 69.00 & 86.09 & 93.27 \\ 
\quad w/ $\alpha$ = 1.5 ~\cite{o2023contrastive} & 81.96 & 97.86 & 89.00 & 65.09 & 69.58 & 86.01 & 93.18  \\
\quad w/ optimal $\alpha$ & 82.18 & \textbf{98.10} & 90.30 & 65.09 & 69.58 & 86.43 & 93.35 \\ 
\quad \quad + entropy sweet spot & \textbf{82.34} & \textbf{98.10} & \textbf{90.50} & \textbf{65.61} & \textbf{70.01} & \textbf{86.60} & \textbf{93.56}\\
\hline
Phi3-medium & 90.83 & 98.33 & 92.60 & 72.81 & 77.00 & 89.42 & 95.33 \\
\midrule
Llama-2-7B & 21.68 & 69.76 & 53.00 & 9.65 & 61.72 & 62.88 & 79.67 \\ 
\quad w/ reference ($\alpha$ = 1) & 22.90 & 67.38 & 54.90 & 9.83 & 61.43 & 63.65 & 80.56 \\ 
\quad w/ $\alpha$ = 1.5 ~\cite{o2023contrastive} & 22.90 & 67.14 & 55.20 & 9.65 & 60.55 & 64.08 & 80.56 \\
\quad w/ optimal $\alpha$ & 23.35 & 70.00 & 55.20 & 10.00 & 62.45 & 64.08 & 80.68 \\ 
\quad \quad + entropy sweet spot & \textbf{24.03} & \textbf{70.48} & \textbf{56.10} & \textbf{10.53} & \textbf{62.59} & \textbf{64.59} & \textbf{80.81} \\ 
\hline
Llama-2-13B & 37.38 & 86.67 & 60.8 & 12.63 & 67.69 & 67.15 & 80.51 \\
\midrule
Vicuna-7B & 19.56 & 61.67 & 45.00 & 7.90 & 65.36 & 62.88 & 80.60 \\ 
\quad w/ reference ($\alpha$ = 1) & 21.08 & 60.71 & 45.80 & 7.90 & 64.77 & 62.29 & 80.77 \\ 
\quad w/ $\alpha$ = 1.5 ~\cite{o2023contrastive} & 20.32 & 60.48 & 46.30 & 7.54 & 65.50 & 62.12 & 80.35 \\
\quad w/ optimal $\alpha$ & 21.08 & 62.38 & 46.60 & 8.42 & 65.65 & 63.91 & 80.98 \\ 
\quad \quad + entropy sweet spot & \textbf{21.38} & \textbf{63.81} & \textbf{46.90} & \textbf{8.77} & \textbf{66.23} & \textbf{64.42} & \textbf{81.36} \\ 
\hline
Vicuna-13B & 36.01 & 85.71 & 55.8 & 12.11 & 66.08 & 69.54 & 83.46 \\
\midrule
Qwen-1.8B (4B on MATH-Easy) & 35.03 & 83.10 & 34.50 & 27.54 & 58.22 & 50.00 & 70.41\\ 
\quad w/ reference ($\alpha$ = 1) & 35.10 & 83.81 & 35.80 & 27.89 & 59.97 & 48.38 & 70.20 \\ 
\quad w/ $\alpha$ = 1.5 ~\cite{o2023contrastive} & 34.72 & 83.57 & 36.30 & 28.42 & 60.12 & 48.21 & 70.12 \\
\quad w/ optimal $\alpha$ & 35.41 & 83.81 & \textbf{37.10} & 28.77 & 60.41 & 50.94 & 70.41 \\ 
\quad \quad + entropy sweet spot & \textbf{36.09} & \textbf{85.95} & \textbf{37.10} & \textbf{29.82} & \textbf{61.14} & \textbf{51.28} & \textbf{71.04} \\ 
\hline
Qwen-4B (7B on MATH-Easy \& StrategyQA) & 46.32 & 91.91 & 60.5 & 30.70 & 67.83 & 65.70 & 78.91 \\
\midrule
\end{tabular}
}
\caption{Experiment results on various LLM benchmarks. The results are in case only the first generated token received knowledge from the teacher model. `entropy sweet spot' is a method using Algorithm~\ref{algorithm_1}.} 
\label{overall_result}
\end{table*}

\begin{figure*}[t]
    \centering
    \includegraphics[width=1.75 \columnwidth]{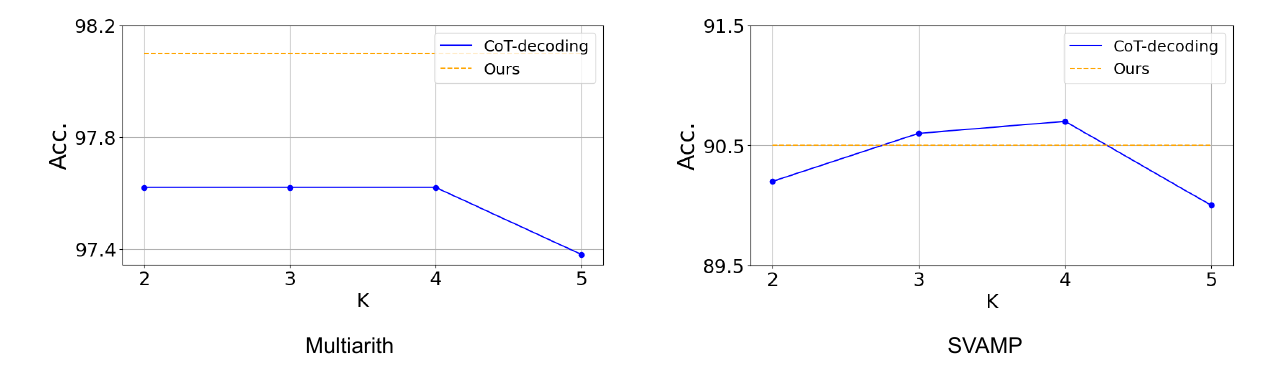}
    
    \caption{Results of comparison of our method with CoT-decoding using the Phi-3. K denotes counts of exploring paths starting from top-$k$.}
    \label{fig:comparision_CoT}
\end{figure*}

We assess experimental results across various classification tasks within diverse domains, as detailed in Table~\ref{result_classification}. Our method consistently outperforms the teacher model across all classification tasks, indicating significant enhancements in model performance.
Moreover, we analyze our method in the MNLI dataset using student models of varied performance and size, illustrated in Table~\ref{mnli_ablation}. The table shows that even if a student model performs significantly worse than the teacher model, an appropriate mixture of their knowledge through an optimal $\alpha$ value can surpass the teacher's performance. 
Additionally, our results show that as the performance gap between the student and teacher narrows, the optimal $\alpha$ value converges to 0.5. This indicates a simple weighted average ensemble, as when there is minimal difference in knowledge between the students and teachers.

\subsection{LLM benchmark}

\paragraph{Overall results.}
We conduct vast experiments to evaluate the effectiveness of our proposed method across various LLMs and diverse benchmarks. The results of these experiments are in Table~\ref{overall_result}. Our approach is applied solely to the first token generated. As demonstrated by the results, directly utilizing the logits from a teacher model and utilizing the previous method~\cite{o2023contrastive} can enhance performance; however, it is slight and occasionally leads to degraded performance. By optimizing the $\alpha$ parameter, we consistently achieve performance improvements over them across all models. Furthermore, we observe a significant enhancement in model performance by determining whether to employ our decoding method based on the entropy value.

\begin{table*}[!]
\small
\centering
\vspace{0.25cm}
\resizebox{1\textwidth}{!}{
\begin{tabular}{lccccccc}
\toprule
 Models & GSM8K & Multiarith & SVAMP & MATH-Easy & StrategyQA & ARC-Challenge & ARC-Easy \\ \hline
Phi3-mini & 80.36 & 97.62 & 89.90 & 63.51 & 67.54 & 85.92 & 92.76\\ 
\quad N = 3 & 82.79 & \textbf{98.10} & 90.10 & \textbf{66.67} & 69.87 & 86.60 & 93.48 \\ 
\quad N = 5 & 82.34 & \textbf{98.10} & 90.10 & 66.32 & 70.31 & \textbf{86.69} & \textbf{93.69} \\ 
\quad N = 10 & \textbf{83.32} & \textbf{98.10} & \textbf{90.30} & \textbf{66.67} & \textbf{72.20} & \textbf{86.69} & 93.43 \\  \midrule

Llama-2-7B & 21.68 & 69.76 & 53.00  & 9.65 & 61.72 & 62.88 & 79.67 \\ 
\quad N = 3 & 23.50 & 70.48 & 56.70 & 9.83 & 63.17 & 63.48 & 80.39 \\ 
\quad N = 5 & 23.43 & \textbf{70.95} & \textbf{56.90} & \textbf{10.18} & 63.46 & \textbf{64.51} & \textbf{80.77} \\ 
\quad N = 10 & \textbf{24.56} & \textbf{70.95} & 56.60 & 10.00 & \textbf{64.05} & 64.42 & 80.56 \\  \midrule

Vicuna-7B & 19.56 & 61.67 & 45.00 & 7.90 & 65.36 & 62.88 & 80.60 \\ 
\quad N = 3 & 20.85 & 63.10 & 45.70 & 9.83 & 66.23 & 64.76 & 81.19\\ 
\quad N = 5 & 20.70 & 64.05 & 46.20 & \textbf{10.00} & \textbf{66.67} & 64.68 & \textbf{81.23} \\ 
\quad N = 10 & \textbf{21.91} & \textbf{64.76} & \textbf{47.00} & 9.30 & 66.52 & \textbf{65.78} & 81.02 \\  \midrule

Qwen-1.8B (4B on MATH-Easy) & 35.03 & 83.10 & 34.50 & 27.54 & 58.22 & 50.00 & 70.41 \\ 
\quad N = 3 & \textbf{35.71} & 83.81 & 39.30 & 29.47 & \textbf{62.45} & 51.45 & 70.79 \\ 
\quad N = 5 & 35.18 & \textbf{84.05} & 41.20 & 30.35 & 61.86 & 52.90 & 70.83 \\ 
\quad N = 10 & 35.63 & 83.57 & \textbf{43.20} & \textbf{31.23} & 61.72 & \textbf{54.18} & \textbf{72.14} \\  \bottomrule

\end{tabular}
}\vspace{-0.5em}
\caption{Experiment results on various LLM benchmarks with diverse N. We use optimal $\alpha$ for all experiments.} \vspace{-1em}
\label{N_ablation_results}
\end{table*}

\begin{table}[t]
\centering
\vspace{0.25cm}
\resizebox{1\linewidth}{!}{
\begin{tabular}{lrrr}
\toprule
\multirow{2}{*}{Methods} 
 & \multicolumn{3}{c}{Seconds / Sentences} \\
 \cmidrule(r){2-4}
 & Total & Phi-3-medium & Phi-3-mini \\
\midrule 
 Student Greedy & 2.52 & - & 2.52 \\
 Teacher Greedy & 5.20 & 5.20 & - \\
 CoT-decoding (K=3) & 8.15 & - & 8.15 \\
 CoT-decoding (K=5) & 14.65 & - & 14.65 \\
 Ours & 2.63 & 0.08 & 2.55 \\
\bottomrule
\end{tabular}
}\vspace{-0.5em}
\caption{Comparative speed analysis of our method against CoT-Decoding. Our approach requires one additional token computation for teacher per sentence.}\vspace{-1em}
\label{Speed_comprasion}
\end{table}

\begin{table}[t]
\centering
\vspace{0.25cm}
\resizebox{1\linewidth}{!}{
\begin{tabular}{lrrr}
\toprule
\multirow{2}{*}{Methods} 
 & \multicolumn{3}{c}{TOTAL FLOPs (TFLOPS)} \\
 \cmidrule(r){2-4}
 & Total & Phi-3-medium & Phi-3-mini \\
\midrule 
 Speculative decoding & 1309.32 & 1074.86 & 234.46 \\
 Ours & 273.52 & 7.65 & 265.87 \\
\bottomrule
\end{tabular}
}\vspace{-0.5em}
\caption{Comparing FLOPs of our method against speculative decoding. For speculative decoding, we set the number of tokens to verify as four.}
\label{FLOPs_comparison}\vspace{-1em}
\end{table}

\paragraph{Comparision with CoT-decoding.} As the concept to reasonably determine the first token, our method is related to CoT-decoding~\cite{wang24}, a methodology that begins by selecting the top-$k$ tokens from the initial token and subsequently generating sentences from each. The most confident response is then chosen as the final output. We compare our method to CoT-decoding using SVAMP and Multiarith, which do not require elaborate prompting, focusing on performance and speed. The performance results are in Fig.~\ref{fig:comparision_CoT}, and the speed comparison is in Table~\ref{Speed_comprasion}. The performance of our method is competitive with CoT-decoding and even surpasses it in Multiarith.
The speed calculated on the A100 GPU shows that our method is faster than CoT-decoding. Time cost in DNN is negligible in that it is smaller than 0.001 seconds per sentence.

\paragraph{Comparision with Speculative decoding.}
Our method shares similarities with speculative decoding in designing an efficient collaboration between small language models and LLMs. However, unlike speculative decoding, we implement a decoding strategy that operates under limited LLM supervision, significantly containing less computational cost. As illustrated in Table~\ref{FLOPs_comparison}, the FLOPs comparison between speculative decoding and our proposed method evidences a substantial reduction in computational costs.

\paragraph{Various tokens setting.} 
We define the extent of collaborative decoding between the student and teacher models by notating `$N$,' the number of tokens for collaboration. Then, we conduct diverse $N$ values to assess the impact on performance, and the results are in Table~\ref{N_ablation_results}. Moreover, we conduct our method on all tokens reported in Appendix~\ref{Dont_Trust_Your_Student}.

\subsection{Optimal $\alpha$ prediction model}

We evaluated our optimal $\alpha$ prediction model using the Phi-3 on the GSM8K, and the results in Table~\ref{alpha_predict_model_result}. While XGBoost failed to classify effectively, the DNN exceeded the performance of the entropy sweet spot. Demonstrates that the model can autonomously predict the appropriate knowledge without needing hand-crafted tuning processes. Additionally, we conducted more experiments for checking generalizability, such as results on unseen benchmarks and joint training, as detailed in Appendix \ref{add_optimal_alpha_exp}.

\begin{table}[t]
\centering
\vspace{0.25cm}
\resizebox{0.8\linewidth}{!}{
\begin{tabular}{lc}
\toprule
 Models & GSM8K \\ \midrule
Phi3-mini  & 80.36 \\ 
\quad w/ entropy sweet spot & 82.34 \\
\quad w/ XGboost & 81.44 {$\pm$ 0.51} \\ 
\quad w/ DNN & \textbf{82.44} {$\pm$ 0.22} \\ 
\bottomrule
\end{tabular}
}\vspace{-0.5em}
\caption{Experiment results on optimal $\alpha$ predictor.} \vspace{-1em}
\label{alpha_predict_model_result}
\end{table}
\begin{table*}[h!]
\centering
\vspace{0.25cm}
\small
\begin{tabu} to \textwidth { X[l] }
        \toprule
        \textbf{[GSM8K]}  \textit{John runs 60 miles a week. He runs 3 days a week.  He runs 3 hours the first day and half as much the other two days he runs.  How fast does he run?} \\
        \textbf{Ground truth answer}: 10 \\
        \textbf{Generate solo}: \textcolor{green}{John} runs 3 hours the first day, so he runs 3/3 = 1 hour per day for the other two days. He runs a total of 3 + 1 + 1 = 5 hours a week. He runs 60 miles a week, so his speed is 60 miles / 5 hours = 12 miles per hour. The answer is \textcolor{red}{12}. \\
        
        \textbf{Receiving knowledge from the teacher (w/ $\alpha$ = 1.5)}:  \textcolor{green}{Let}'s break this down step by step.

1. John runs 3 hours on the first day.

2. He runs half as much time on the other two days, so that's 3 / 2 = 1.5 hours per day.

3. In total, he runs for 3 + 1.5 + 1.5 = 6 hours a week.

4. He runs 60 miles a week, so to find his speed, we divide the total distance by the total time: 60 miles / 6 hours = 10 miles per hour.

John runs at a speed of 10 miles per hour. The answer is \textcolor{blue}{10}.
\\

        \textbf{Receiving knowledge from the teacher (w/ optimal $\alpha$)}: \textcolor{green}{Let}'s break this down step by step.

1. John runs 3 hours on the first day.

2. He runs half as much time on the other two days, so that's 3 / 2 = 1.5 hours per day.

3. In total, he runs for 3 + 1.5 + 1.5 = 6 hours a week.

4. He runs 60 miles a week, so to find his speed, we divide the total distance by the total time: 60 miles / 6 hours = 10 miles per hour.

John runs at a speed of 10 miles per hour. The answer is \textcolor{blue}{10}.
\\

        \textbf{Teacher generation}: \textcolor{green}{First}, we need to find out how many hours John runs in total. He runs 3 hours the first day and half as much the other two days. So, he runs 3 + (3/2) + (3/2) = 3 + 1.5 + 1.5 = 6 hours in total. Now, we know he runs 60 miles in 6 hours. To find his speed, we divide the distance by the time. So, he runs 60 miles / 6 hours = 10 miles per hour. The answer is \textcolor{blue}{10}.
        \\

        \midrule
        
        \textbf{[StrategyQA]}  \textit{Is Argon near Neon on the periodic table of elements?} \\
        \textbf{Ground truth answer}: True (yes) \\
        \textbf{Generate solo}: \textcolor{green}{Argon} is in period 3 and Neon is in period 2. Periods are horizontal rows on the periodic table. Thus, Argon is not near Neon on the periodic table. So the answer is \textcolor{red}{no}. \\

        \textbf{Receiving knowledge from the teacher (w/ $\alpha$ = 1.5)}: \textcolor{green}{Argon} is in period 3 and Neon is in period 2. Periods are horizontal rows on the periodic table. Thus, Argon is not near Neon on the periodic table. So the answer is \textcolor{red}{no}. \\
        
        \textbf{Receiving knowledge from the teacher (w/ optimal $\alpha$)}: \textcolor{green}{Yes}. Argon is the 18th element on the periodic table. Neon is the 10th element on the periodic table. Thus, Argon is near Neon on the periodic table of elements. So the answer is \textcolor{blue}{yes}. 
        \\

        \textbf{Teacher generation}: \textcolor{green}{Argon} and Neon are both noble gases. Noble gases are in the same column on the periodic table. Thus, Argon is near Neon on the periodic table. So the answer is \textcolor{blue}{yes}. \\
         \bottomrule
\end{tabu}
\vspace{-0.5em}
\caption{Detailed results showing how our method brought performance gain in LLM benchmarks utilizing Phi-3 model. Each generated tokens all in green; if the answer is correct we highlighted in blue, else red. }
\label{tab:case_studys}\vspace{-1em}
\end{table*}

\subsection{Case study}

Table~\ref{tab:case_studys} provides example cases where only the first generated token receives knowledge from a teacher. As follows, receiving initial guidance from the teacher encourages the student to generate correct solutions. A detailed analysis of the first sample from the GSM8K dataset reveals that `Receiving knowledge from the teacher’ facilitates the subsequent student generation process by dramatically deviating from the initial tokens of both the student and the teacher. Notably, in the second example, our method begins by generating the correct answer “Yes,” reflecting the fact that the teacher has a much higher probability of generating “Yes” than the student; this results in better conditioning of the subsequent tokens, leading to correct reasoning as a result. This example suggests why an $\alpha$ of 1.5 is not optimal, commonly used in previous research~\cite{o2023contrastive}.



\section{Related work}\label{sec:related}

\paragraph{Efficient decoding strategy.}
Speculative decoding \cite{xia23, leviathan23, zhou2024distillspec} enables faster generation by having the draft model generate first, which the target model then verifies in parallel, ensuring distributions identical to those generated by the target model. On the other hand, our method aims to efficiently enhance the student model's performance by utilizing only a few tokens generated by the teacher.

\paragraph{Logit arithmetic.}
Recent works suggest using arithmetic to harness the capabilities of two or more language models during decoding.
Contrastive decoding~\cite{li-etal-2023-contrastive, o2023contrastive} enhances LLM generation quality 
by subtracting the log probabilities of a smaller model from those of the LLM.
SafeDecoding~\cite{xu2024safedecoding} mitigates jailbreak attacks by amplifying the probabilities of tokens aligned with human values, using the distribution difference between the safety-enhanced expert model and the original model.
Other analogous studies utilize differences in logits over vocabulary distributions between sLLMs to influence the output of LLM.
DExperts~\cite{liu21} suggests a decoding strategy to reduce undesirable outputs of target LLM by leveraging "expert" LMs and "anti-experts" LMs. Using a similar equation, Emulator fine-tuning~\cite{mitchell2024emulator} emulates instruction-tuning large pre-trained models by ensembling them with small fine-tuned models. \cite{zhao2024weak} uses a small unsafe model to manipulate the output distribution of a larger model to induce jailbreaking. 
Finally, proxy-tuning \cite{liu24} utilizes the difference in output logits between tuned and untuned small LMs to shift the predictions of the untuned large LMs. 
Unlike previous methods, our approach involves searching for the optimal mixing coefficient $\alpha$ across a broader range. 
We discover that the optimal point may differ from those used in previous studies. Our work also employs an entropy-based method and a model-driven approach to enhance the general reasoning ability of LMs.

\paragraph{Importance of the first token.} The chain-of-thought decoding~\cite{wang24} shows that the first token that LLMs produce have a significant impact on the quality of the entire response in the reasoning task. Exploring paths starting from top-$k$ alternative tokens enables chain-of-thought reasoning without prompts. Similarly, our study also shows that variations in the initial token can greatly impact the model's performance. However, our method only requires a
single token from the teacher model,
eliminating the need for additional computation to produce all possible $k$ outcomes.

\section{Conclusion}\label{sec:concl}
In this paper, we have formulated the problem of sLLM decoding with a limited LLM supervision. Through our study, we have unveiled that the optimal combination of sLLM and LLM predictions may significantly depend on the considered task, models, and even datum. Understanding when and why such discrepancies happen is an important future question that needs to be addressed. We believe that our entropy-based perspective will help provide a strong starting point for this pursuit.



\section*{Limitations}
A notable limitation of our method is that, in its current form, it is difficult to incorporate the predictions of the teacher that has a different embedding space, \eg, using a different tokenizer. Another limitation is that we rely on a single feature, namely the prediction entropy, to determine how to aggregate the predictions. A more in-depth analysis of what other features one can utilize is much needed.


\section*{Ethics statement}
All experimental results we provide in this paper is based on publicly available datasets and open-source models.

\section*{Acknowledgments}
This research was supported in part by the NAVER-Intel Co-Lab, and in part by the National Research Foundation of Korea (NRF) grant funded by the Korea government (MSIT) (No. RS-2024-00453301). The work was conducted by POSTECH and reviewed by both NAVER and Intel.

\bibliography{custom}

\newpage
\clearpage

\appendix
\section{Detailed empirical study in optimal $\alpha$ cases} \label{detailed empirical study in entropy}
\begin{figure}[t]
    \centering
    \includegraphics[width=1 \columnwidth]{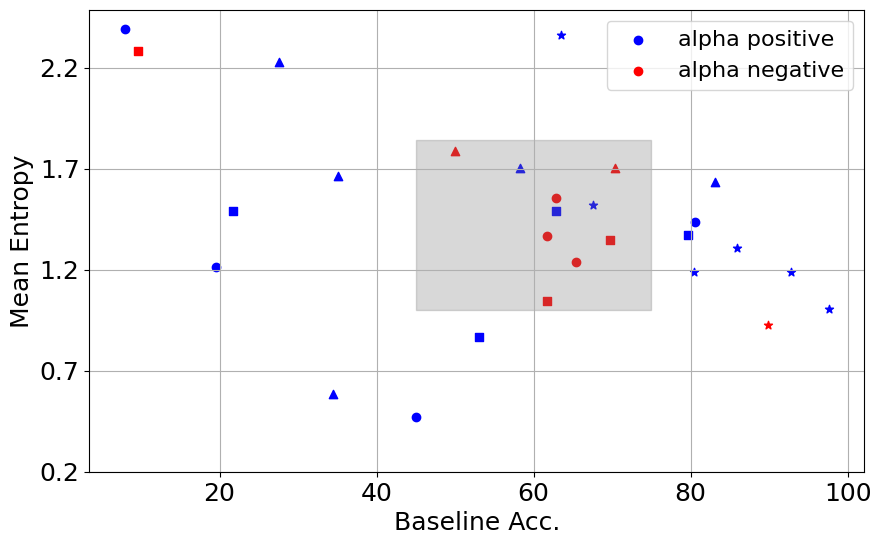}
    
    \caption{Illustration of the correlation between performance, mean entropy, and the sign of optimal $\alpha$. Different geometric shapes represent distinct models.}
    \label{fig:optimal_alpha_case}
\end{figure}

We explore the relationship between the signs of the $\alpha$ (positive/negative) and their baseline performance and entropy of the first token, as depicted in Fig.~\ref{fig:optimal_alpha_case}. As illustrated, the $\alpha$ values are predominantly positive when the baseline accuracy and entropy are either high or low. However, $\alpha$ values in the intermediate performance and entropy, which in the gray zone of Fig.~\ref{fig:optimal_alpha_case} generally exhibit negative values—three instances where positive $\alpha$ values in the gray zone own significant performance disparities between the teacher and student models. Furthermore, our observations reveal that high entropy does not always correlate with performance, suggesting that while high entropy can indicate challenging problem instances, it may also reflect considerate thinking by the model, and teachers' knowledge can be confused. These findings support our entropy-based Algorithm~\ref{algorithm_1}.

\section{Detailed informations}\label{app:details}

To demonstrate the superiority of our method, we conduct experiments with various models and datasets. The details of the model, hyper-parameter settings, and datasets are below.

\begin{table}[h]
\centering
\resizebox{\linewidth}{!}{
\begin{tabular}{lcccc}
\toprule
\multirow{2}{*}{Datasets} 
 & \multicolumn{3}{c}{Hyper-parameters} \\
\cmidrule(r){2-4}
& epochs & batch size & lr \\
\midrule
CIFAR-100 & 20 & 128 & $5 \times 10^{-4}$ \\
ESC-50 & 10 & 32 & $1 \times 10^{-4}$ \\
MNLI Teacher model$^\dag$ & 3 & 64 & $3 \times 10^{-6}$ \\
MNLI Student model$^\dag$ & 3 & 64 & $1 \times 10^{-5}$ \\
\bottomrule
\end{tabular}
}
\caption{Hyper-parameter settings on classification task. The same parameter settings are used for both students and teachers. The parameter setting marked with $^\dag$ are from~\citet{he2020deberta} }
\label{tab:detail classification parameter}
\end{table}
\begin{table}[h]
\centering
\resizebox{\linewidth}{!}{
\begin{tabular}{lcccc}
\toprule
\multirow{2}{*}{Models} 
 & \multicolumn{4}{c}{Hyper-parameters} \\
\cmidrule(r){2-5}
& epochs & batch size & lr & max depth \\
\midrule
XGboost & - & - & $3 \times 10^{-1}$ & 6 \\
DNN & 5 & 1024 & $5 \times 10^{-7}$ & - \\
\bottomrule
\end{tabular}
}
\caption{Hyper-parameter settings on optimal $\alpha$ predict model. }
\label{tab:detail alpha model parameter}
\end{table}
\begin{table*}[!]
\small
\centering
\vspace{0.25cm}
\resizebox{0.8\textwidth}{!}{
\begin{tabular}{lcc}
\toprule
 & \multicolumn{2}{c}{\multirow{2}{*}{\textbf{Classification task}}}\\
\\
\cmidrule(r){2-3}
 Task & Student & Teacher \\ \midrule
CIFAR-100 & facebook/deit-base-patch16-224 & facebook/deit-tiny-patch16-224 \\ 
ESC-50 & Only three layers of the Teacher & MIT/ast-finetuned-audioset-16-16-0.442 \\ 
MNLI & microsoft/deberta-large-mnli & microsoft/deberta-v2-xxlarge-mnli \\ 
\midrule
\midrule
 & \multicolumn{2}{c}{\multirow{2}{*}{\textbf{LLM benchmarks}}}\\
\\
\cmidrule(r){2-3}
 Models & Student & Teacher \\ \midrule
Llama-2 & meta-llama/Llama-2-7b-chat-hf & meta-llama/Llama-2-13b-chat-hf \\ 
Vicuna-v1.5 & lmsys/vicuna-7b-v1.5 & lmsys/vicuna-13b-v1.5 \\ 
Qwen1.5 & Qwen/Qwen1.5-1.8B-Chat & Qwen/Qwen1.5-4B-Chat \\ 
Phi-3 & microsoft/Phi-3-mini-4k-instruct & microsoft/Phi-3-medium-4k-instruct\\ 
\bottomrule

\end{tabular}
}
\caption{Teacher \& Student model information.} 
\label{tab:detail model info}
\end{table*}

\begin{table}[t]
\centering
\resizebox{\linewidth}{!}{
\begin{tabular}{lccccc}
\toprule
\multicolumn{5}{c}{\multirow{2}{*}{\textbf{Classification tasks}}}\\
\\
\midrule
Datasets & \#Total & \#Train & \#Dev & \#Test     \\
\midrule
CIFAR-100 & 60,000 & 50,000 & - & 10,000    \\
ESC-50 & 2,000 & 2,000 & - & -     \\
MNLI-matched & 431,993  & 392,702 & 9,815 & 9,796     \\
MNLI-mismatched & - & - & 9,832 & 9,847     \\
\midrule
\midrule
\multicolumn{5}{c}{\multirow{2}{*}{\textbf{LLM benchmarks}}}\\
\\
\midrule
Datasets & \#Total & \#Train & \#Dev & \#Test     \\
\midrule
GSM8K & 8,792 & 7,473 & - & 1,319    \\
Multiarith & 600 & 420 & - & 180    \\
SVAMP & 1,000 & 700 & - & 300    \\
MATH-easy & 966 & 570 & - & 396    \\
StrategyQA & 2,290 & 1,603 & - & 687    \\
ARC-challenge & 2,590 & 1,119 & 299 & 1,172   \\
ARC-easy & 5,197 & 2,251 & 570 & 2,376    \\
\bottomrule
\end{tabular}
}
\caption{Dataset statistics information.}
\label{Dataset Statistics}
\end{table}

\subsection{Model information}
\label{detail model}
The detailed model information for students and teachers is in Table~\ref{tab:detail model info}. In the exceptional case of StrategyQA, we utilize the Qwen model with a 7B model as the teacher reasons that the 4B model exhibited inferior performance compared to the 1.8B model. On MATH-easy, as the accuracy is 0 in the 1.8B Qwen model, we utilize student a 4B model and 7B for the teacher. For the optimal $\alpha$ prediction model, we make a simple 5-layer DNN.

\subsection{Hyper-parameter settings}
\label{detail hyper-parameter}

We conduct our experiments with hyper-parameter settings as outlined in Table~\ref{tab:detail classification parameter} for classification and Table~\ref{tab:detail alpha model parameter} optimal $\alpha$ prediction model. We utilize AdamW~\cite{loshchilov2018decoupled} optimizer for classification task and for optimal $\alpha$ predict DNN. We performed a grid search for the learning rate, batch size, and weight decay to find an optimal value.

\subsection{Dataset statistics}
\label{detail data information}

We conduct experiments on a range of datasets to demonstrate the superiority of our method for various domain classification tasks and LLM benchmarks. The overall dataset statistics are illustrated in Table~\ref{Dataset Statistics}, and each task description is below.

\subsubsection{Classification task}
\noindent\textbf{CIFAR-100.}  
Popular datasets for image classification. It consists of 32 × 32-color images, 50,000 for the train set and 10,000 for the test set with 100 classes.

\noindent\textbf{ESC-50.}  
The audio classification dataset comprises 2,000 environmental sound recordings, each lasting 5 seconds. These recordings are annotated across 50 classes, each allocated a single class label. 

\noindent\textbf{MNLI.}  
The Nature Language Inference dataset is part of the GLUE benchmark~\cite{wang2018glue}. It consists of 431,993 data with three classes, and the test set consists of two distinct parts. The `matched' is from the same source as the training dataset, and the `mismatched' is from a different domain.

\subsubsection{LLM benchmarks}  
\label{LLM_benchmark_details}
\noindent\textbf{GSM8K}. 
A dataset of 8,792 linguistically diverse math word problems for grade school students, crafted by human problem writers \cite{cobbe2021training}. It is intended for multi-step mathematical reasoning, with problems requiring between 2 to 8 steps to solve. Solutions mainly involve performing a series of basic arithmetic operations to arrive at the final answer. We use test set for evaluation in this paper.

\noindent\textbf{Multiarith}. 
A collection of multi-step arithmetic problems from MAWPS \cite{koncel-kedziorski-etal-2016-mawps}. These problems require several reasoning steps involving simple arithmetic calculations to obtain the final answer. Because of the low resource test data, we use train set.

\noindent\textbf{SVAMP}.
A challenge set for elementary-level math word problems. Each problem consists of a brief natural language narrative that describes a scenario and asks a question about certain unknown quantities. We use both train and test set for evaluation.

\noindent\textbf{MATH}. 
Dataset that consists of 12,500 problems from high school math competitions \cite{hendrycks2021measuring}. The problems are challenging because they require a higher level of step-by-step mathematical problem-solving skills to solve. Cause many models have poor performance, we predicted some easy tasks we defined as MATH-Easy, which contains the train set with level 1 and 3 algebra parts. 

\noindent\textbf{StrategyQA} focuses on open-domain 2,780 questions that require models to deduce a multi-hop strategy to answer. We use test set in this paper.

\noindent\textbf{ARC}. 
The AI2’s Reasoning Challenge (ARC) dataset designed for multiple-choice question-answering, encompasses science exam questions ranging from grades 3 to 9. It is partitioned into Easy and Challenge sections, with the latter comprising more complex questions that demand reasoning skills.  We use test set in this paper.

\section{Additional results}
\subsection{More experiments of optimal $\alpha$ predict model}
\label{add_optimal_alpha_exp}
Through additional empirical analysis, we find that alpha prediction models are generalizable, at least to some extent.
We perform two sets of experiments. As shown in Table~\ref{alpha_predict_model_result_add_exp1}, we trained the alpha predictor on GSM8K and evaluated it on Multiarith and MATH-Easy (all math datasets). We observe that the predictor achieves a close performance to the “entropy sweet spot,” demonstrating effective generalizability to unseen data. As illustrated in Table~\ref{alpha_predict_model_result_add_exp2}, we trained the alpha predictor on a joint set of GSM8K, ARC-Challenge, and ARC-Easy and predicted on each dataset. Again, we observe that the predictor performs well on each task, indicating that one can train an adequate alpha predictor by combining data from various sources.

\begin{table}[t]
\centering
\vspace{0.25cm}
\resizebox{1\linewidth}{!}{
\begin{tabular}{lcc}
\toprule
 Models & Multiarith & MATH-Easy \\ \midrule
Phi3-mini & 97.62 & 63.51 \\ 
\quad w/ reference ($\alpha$ = 1) & 97.62 & 64.39 \\
\quad w/ optimal $\alpha$ & 98.10 & 65.09 \\
\quad w/ entropy sweet spot & 98.10 & 65.61 \\
\quad w/ DNN (Unseen) & 97.95 {$\pm$ 0.13} & 65.86 {$\pm$ 0.50} \\ 
\bottomrule
\end{tabular}
}
\caption{Experiment results on unseen benchmarks. We train with GSM8K and evaluate with Multiarith and MATH-Easy.} 
\label{alpha_predict_model_result_add_exp1}
\end{table}

\begin{table}[t]
\centering
\vspace{0.25cm}
\resizebox{1\linewidth}{!}{
\begin{tabular}{lccc}
\toprule
 Models & GSM8K & ARC-Challenge & ARC-Easy \\ \midrule
Phi3-mini  & 80.36 & 85.92 & 92.76 \\ 
\quad w/ reference ($\alpha$ = 1) & 81.50 & 86.09 & 93.27 \\
\quad w/ optimal $\alpha$ & 82.18 & 86.43 & 93.35 \\
\quad w/ entropy sweet spot & 82.34 & 86.60 & 93.56 \\
\quad w/ DNN & 81.91 {$\pm$ 0.20} & 86.47 {$\pm$ 0.20} & 93.29 {$\pm$ 0.09} \\ 
\bottomrule
\end{tabular}
}
\caption{Experiment results on joint training. We trained with the GSM8K, ARC-Challenge, and ARC-Easy train set and evaluated on each benchmark's test set.} 
\label{alpha_predict_model_result_add_exp2}
\end{table}

\subsection{The rank of a new word in the original distribution }
\begin{figure*}[t]
    \centering
    \includegraphics[width=2 \columnwidth]{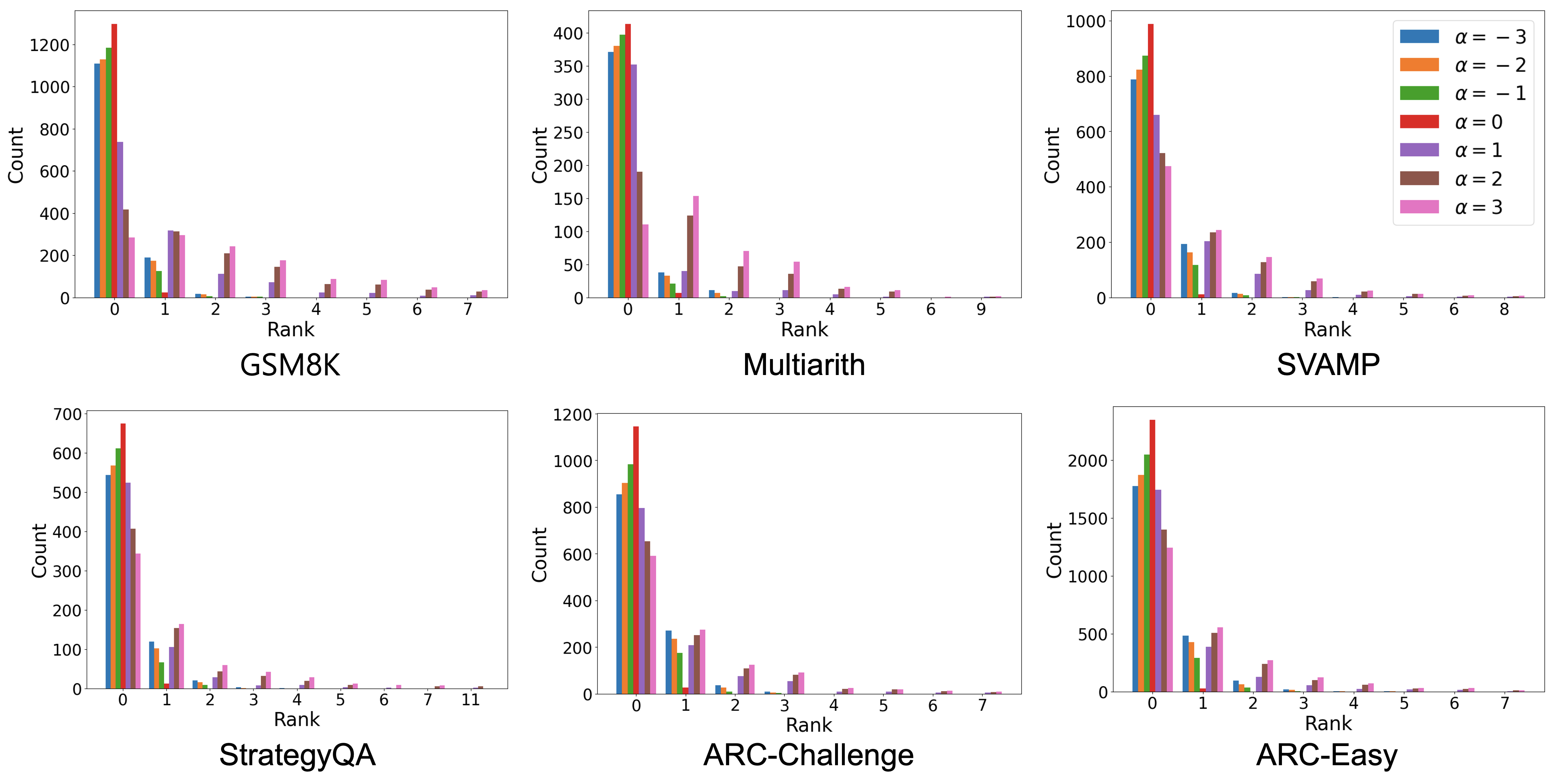}
    \caption{The frequency of rankings of new words in the evaluation of the Phi3-mini. }
    \label{fig:phi3-mini_rank}
\end{figure*}
Fig.~\ref{fig:phi3-mini_rank} shows the frequency of the ranks of new tokens generated with supervision from the teacher model within the distribution of tokens predicted by the student model, according to $\alpha$. As evident from the results, as $\alpha$ moves away from zero, our algorithm selects more low-ranked words. This indicates that effectively selecting new words that lead to coherent sentences through extrapolation may result in dramatic performance gains.

\begin{table}[t]
\centering
\vspace{0.25cm}
\resizebox{0.7\linewidth}{!}{
\begin{tabular}{lc}
\toprule
 Models & GSM8K \\ \hline
Phi3-mini (Student)  & 80.36 \\ 
Phi3-medium (Teacher)  & 90.83 \\ 
DTYS & \textbf{91.51} \\
\midrule
Llama2-7B (Student) & 21.68 \\ 
Llama2-13B (Teacher) & 37.38 \\ 
DTYS & \textbf{38.74} \\
\midrule
Vicuna-7B (Student) & 19.56 \\ 
Vicuna-13B (Teacher) & 36.01 \\ 
DTYS & \textbf{37.53} \\
\midrule
Qwen-1.8B (Student) & 35.03 \\ 
Qwen-4B (Teacher) & 46.32 \\ 
DTYS & \textbf{46.78} \\
\bottomrule
\end{tabular}
}
\caption{Experiment results on Don't trust your student.} 
\label{DTYS}
\end{table}

\subsection{Don't Trust Your Student}
\label{Dont_Trust_Your_Student}

Based on the transformation of equation~\ref{equation1} into equation~\ref{equation2}, extending our decoding method between the student and teacher models to all tokens and setting $\alpha$ higher than 1 can be considered deducting the student's logits from the teachers'. This operation disregards the knowledge represented solely by the student's logits, not trusting the student's knowledge. We term this methodology `Don't Trust Your Student'(DTYS) and apply it to GSM8K. The results, as shown in Table~\ref{DTYS}, indicate an overall performance improvement over the teacher model baseline, demonstrating that DTYS is an effective strategy for leveraging a less capable model to enhance the performance of a more qualified one.

\begin{equation}
\label{equation2}
S_{\alpha}' = \alpha\cdot \sigma(f_t(x)) - (\alpha -1)\cdot \sigma(f_s(x))
\end{equation}
\section{Full Prompts of each benchmark} 
\label{full prompts}
For the GSM8K, Multiarith, SVAMP, and StrategyQA, we refer to the chain-of-thought prompt used in \cite{wei2022chain}. We use 2 shots for the GSM8K, 6 shots for StrategyQA, and 8 shots for the rest.
The full prompts used in the MATH and ARC benchmarks are shown in Table~\ref{prompt_math} and Table~\ref{prompt_arc}, respectively. Due to space constraints in the paper, we omit line breaks between each sentence in MATH and between questions and options in ARC.

\begin{table*}[h]
\centering
\vspace{0.25cm}
\small
\begin{tabular}{p{\textwidth}}
\toprule
\textbf{Prompt for MATH} \\ \midrule
\begin{lstlisting}[breaklines=true, breakatwhitespace=true, keepspaces=true]
Question: What is the largest value of $x$ such that the expression \[\dfrac{x+1}{8x^2-65x+8}\] is not defined?
Let's think step by step. In this particular case, the fraction will be undefined only if its denominator is equal to zero. Because of this, we can ignore the numerator. We start by setting the binomial in the denominator equal to 0: \begin{align*} 8x^2-65x+8=0 \\\Rightarrow\qquad (8x-1)(x-8)=0 \end{align*} We find that the two possible values for $x$ are $\frac18$ and $8$. Since the question asks for the largest value, the final solution is $\boxed{8}$. The answer is 8
Question: BoatWorks built 3 canoes in January of this year and then each subsequent calendar month they built twice the number of canoes they had built the previous month. How many total canoes were built by BoatWorks by the end of March of this year?
Let's think step by step. The number of boats built is $3+3\cdot2+3\cdot2^2 = 3+6+12 = \boxed{21}$. The answer is 21
Question: If $5a+2b=0$ and $a$ is two less than $b$, what is $7b$?
Let's think step by step. First we begin by solving the system of equations \begin{align*} 5a+2b&=0, \\ b-2&=a. \end{align*} Making the substitution for $a$ from the second equation to the first, we get $5(b-2)+2b=0$, which simplifies to $7b-10=0$. Solving for $b$, we find that $b=\frac{10}{7}$. Hence $7b=7\cdot \frac{10}{7}=\boxed{10}$. The answer is 10
Question: The difference between two numbers is 9, and the sum of the squares of each number is 153. What is the value of the product of the two numbers?
Let's think step by step. Call the first number $x$ and the second number $y$. Without loss of generality, assume $x > y$. We can represent the information given in the problem with the following system of linear equations: \begin{align*} x - y &= 9\\ x^2 + y^2 &= 153 \end{align*} Solving for $x$ in the first equation and substituting into the second yields $(9+y)^2 + y^2 = 153$, or $2y^2 + 18y - 72 = 0$. Canceling a $2$ gives $y^2 + 9y - 36 = 0$, which factors into $(y+12)(y-3)$. Thus, $y = 3$ and $x = 12$. So, $x \cdot y = \boxed{36}$. The answer is 36
Question: Simplify $\frac{1}{1+\sqrt{2}}\cdot\frac{1}{1-\sqrt{2}}$.
Let's think step by step. Multiplying the numerators simply yields $1$. Multiplying the denominators gives $1+\sqrt{2} - \sqrt{2} -2 = 1 - 2 = -1$. So, the answer is $\frac{1}{-1} = \boxed{-1}$. The answer is -1
Question: Four people can mow a lawn in 6 hours. How many more people will be needed to mow the lawn in 4 hours, assuming each person mows at the same rate?
Let's think step by step. The number of people mowing and the time required to mow are inversely proportional. Letting $n$ be the number of people and $t$ be the amount of time, we have $nt = (4)(6)= 24$ because 4 people can mow a lawn in 6 hours. If $m$ people can mow the lawn in 4 hours, then we must have $m(4) = 24$, so $m=6$. Therefore, we need $6-4 = \boxed{2}$ more people to complete the job in 4 hours. The answer is 2
Question: BoatsRUs built 7 canoes in January of this year and then each subsequent calendar month they built twice the number of canoes they had built the previous month. How many total canoes were built by BoatsRUs by the end of May of this year?
Let's think step by step. The numbers of canoes built by BoatsRUs each month form a geometric sequence: 7, 14, 28, 56, 112. The first term is 7 and the common ratio is 2, so the sum of these terms is $\frac{7(2^5-1)}{2-1} = \boxed{217}$. The answer is 217
Question: Find the coefficient of the $x^2$ term in the expansion of the product $(ax^3 + 3x^2 - 2x)(bx^2 - 7x - 4)$. 
Let's think step by step. We only need to worry about the terms that multiply to have a degree of $2$. This would be given by the product of the terms $3x^2$ and $-4$ as well as the product of the terms $-2x$ and $-7x$. Thus, $$(3x^2) \times (-4) + (-2x) \times (-7x) = -12x^2 + 14x^2 = 2x^2,$$and the coefficient is $\boxed{2}$. The answer is 2
\end{lstlisting}
\\ \bottomrule
\end{tabular}
\caption{Few-shot exemplars for full chain of thought prompt for MATH}
\label{prompt_math}
\end{table*}

\begin{table*}[h]
\centering
\vspace{0.25cm}
\begin{tabu} to \textwidth { X[l] }
\toprule
\textbf{Prompt for ARC}   \\ \midrule
        
\textbf{Q}: George wants to warm his hands quickly by rubbing them. Which skin surface will produce the most heat?
(A) dry palms (B) wet palms (C) palms covered with oil (D) palms covered with lotion \\
\textbf{A}: Let's think step by step. 
Friction produces heat. Dry skin has a higher coefficient of friction than wet or oily skin. Therefore, dry palms will produce the most heat when rubbed together. 
Therefore, the answer is (A).

\textbf{Q}: Which of the following statements best explains why magnets usually stick to a refrigerator door? 
(A) The refrigerator door is smooth. (B) The refrigerator door contains iron. (C) The refrigerator door is a good conductor. (D) The refrigerator door has electric wires in it.\\
\textbf{A}: Let's think step by step. 
Magnets are attracted to ferrous materials, which contain iron. The refrigerator door is usually made of steel, which is an alloy of iron and carbon. The presence of iron in the refrigerator door allows the magnet to stick to it.\\
Therefore, the answer is (B).

\textbf{Q}: A fold observed in layers of sedimentary rock most likely resulted from the 
(A) cooling of flowing magma. (B) converging of crustal plates. (C) deposition of river sediments. (D) solution of carbonate minerals.\\
\textbf{A}: Let's think step by step. 
Sedimentary rocks are formed from the accumulation and compaction of sediments, such as sand, mud, and organic matter. 
When these sediments are deposited, they are typically laid down in horizontal layers. 
If these layers are later subjected to compressional forces, such as those that occur when two crustal plates converge, they can be folded. 
The folding of the layers is a result of the deformation of the rock under the influence of stress. 
The other options (A, C, and D) are less likely to cause folding in sedimentary rocks. 
Cooling of flowing magma (A) can cause the formation of igneous rocks, but it does not typically cause folding in sedimentary rocks. 
Deposition of river sediments (C) can result in the formation of sedimentary layers, but it does not typically cause folding. 
Solution of carbonate minerals (D) can cause the formation of caves and other karst features, but it does not typically cause folding in sedimentary rocks.
Therefore, The answer is (B).

\textbf{Q}: Which of these do scientists offer as the most recent explanation as to why many plants and animals died out at the end of the Mesozoic era? 
(A) worldwide disease (B) global mountain building (C) rise of mammals that preyed upon plants and animals (D) impact of an asteroid created dust that blocked the sunlight \\
\textbf{A}: Let's think step by step. 
The question asks for the most recent explanation for the mass extinction at the end of the Mesozoic era. 
Answer choice (A), worldwide disease, is not a widely accepted explanation for the mass extinction. 
Answer choice (B), global mountain building, is not a widely accepted explanation for the mass extinction. 
Answer choice (C), rise of mammals that preyed upon plants and animals, is not a widely accepted explanation for the mass extinction. 
Answer choice (D), impact of an asteroid created dust that blocked the sunlight, is the most widely accepted explanation for the mass extinction. 
Therefore, the answer is (D).

\textbf{Q}: A boat is acted on by a river current flowing north and by wind blowing on its sails. The boat travels northeast. In which direction is the wind most likely applying force to the sails of the boat? 
(A) west (B) east (C) north (D) south \\
\textbf{A}: Let's think step by step. 
The boat is traveling northeast, which means it is moving both north and east.
The river current is flowing north, so it is not causing the boat to move east. 
Therefore, the wind must be blowing on the sails of the boat in a direction that causes the boat to move east. 
The only option that satisfies this requirement is (B) east.
Therefore, the answer is (B).
 
 \\ \bottomrule
\end{tabu}
\caption{Few-shot exemplars for full chain of thought prompt for ARC.}
\label{prompt_arc}
\end{table*}

\end{document}